\documentclass[10pt]{article} 
\usepackage[preprint]{tmlr}

\usepackage{amsmath,amsfonts,bm}

\def\eqref#1{equation~\ref{#1}}

\def\1{\bm{1}}

\DeclareMathAlphabet{\mathsfit}{\encodingdefault}{\sfdefault}{m}{sl}
\SetMathAlphabet{\mathsfit}{bold}{\encodingdefault}{\sfdefault}{bx}{n}

\usepackage{hyperref}
\usepackage{url}
\usepackage{booktabs}
\usepackage{lscape}
\usepackage{graphicx}
\usepackage[table]{xcolor}
\usepackage{tikz}
\usetikzlibrary{arrows.meta,positioning,calc,fit,backgrounds}
\definecolor{lightgreen}{rgb}{0.88, 0.95, 0.88}

\title{Closing the Context Gap: Activation Alignment for Tabular In-Context Learning}

\author{\name Yoel Zeldes \email yoel.zeldes@gmail.com \\
      \addr Independent Researcher}

\def\month{MM}  
\def\year{YYYY} 
\def\openreview{\url{https://openreview.net/forum?id=XXXX}} 

\begin{document}

\maketitle

\begin{abstract}
Tabular foundation models perform in-context learning (ICL) by conditioning predictions on labeled training examples provided as context. Unlike traditional models that separate training from inference, these models must process all training examples in every forward pass, making each prediction expensive. Restricting the number of training examples reduces this cost but substantially degrades performance. Instead of discarding context, we propose \emph{activation alignment}, a method that leverages the full context to teach a model how to behave when seeing only a subset. This is achieved by training a lightweight linear transformation on synthetic unlabeled data to map the intermediate activations of a data-constrained ``student'' (using partial context) toward those of a full-context ``teacher'' (using all data). Training the aligner requires no GPU and converges in seconds to minutes on commodity hardware. We evaluate on 38 classification datasets from the TabArena benchmark using the leading two tabular foundation models, TabPFN-3 and TabFM. Across all context budgets, the aligned student yields broad, statistically significant improvements over the unaligned baseline for both models. In low-data regimes, alignment recovers nearly half of the teacher's predictive advantage. The method provides a practical, low-overhead approach to achieving the inference speed of compact contexts while closing a significant fraction of the performance gap to the full-context teacher.
\end{abstract}

\section{Introduction}
\label{sec:introduction}

Tabular foundation models have become the state of the art for structured data prediction \citep{hollmann2025accurate}. Models such as TabPFN-3 \citep{grinsztajn2026tabpfn3technicalreport} and TabFM \citep{tabfm2026} perform in-context learning (ICL): given a set of labeled training examples as context and an unlabeled query, they produce a prediction in a single forward pass without weight updates. This paradigm enables rapid adaptation to new datasets and eliminates the need for task-specific training.

However, the predictive quality of ICL depends heavily on the number of context examples. When the context is small (due to limited data availability, hardware memory constraints, or computational cost of the quadratic attention mechanism), performance degrades substantially. \citet{garg2022transformers} showed that transformers implementing ICL improve with the number of in-context examples, and tabular foundation models exhibit a similar dependence on context size in practice \citep{qu2025tabicl, ma2024tabdpt}, making sample efficiency a central bottleneck for practical deployment.

The root cause is architectural: unlike tree-based models, where training and inference are decoupled and prediction is immediate, tabular foundation models must carry all training examples through the forward pass at every prediction. Although intermediate key--value states could be cached, doing so requires large device memory. Existing approaches reduce this inference cost by compressing the context into compact latent tokens (TACO; \citealp{zabergja2025taco}) or by pruning redundant transformer layers (TACTICL; \citealp{koshil2025tacticl}). Both methods are effective but demand substantial GPU training.

Inspired by prior work on activation steering in large language models \citep{turner2023steering, li2023inference, todd2024function} and feature-based knowledge distillation \citep{romero2015fitnets}, we take a complementary approach. Rather than compressing the input or the architecture, we ask: can a model that has seen the full context teach a data-constrained copy of itself how to behave, by correcting its internal representations?

In this work, we propose \textbf{activation alignment}, a framework that fits a lightweight linear module to project the intermediate activations of a data-constrained student model toward those of a full-context teacher. Similar to TACO, the aligned student operates with a reduced number of input tokens compared to the full dataset. However, the aligner is drastically cheaper to train: a single linear layer fitted on pre-extracted activations, requiring no GPU and converging within seconds to minutes on commodity hardware. At inference time, the aligner corrects the student's hidden states at a single transformer layer without requiring any model fine-tuning.

We systematically evaluate activation alignment across 38 classification tasks from the TabArena benchmark \citep{erickson2025tabarena}, a standardized evaluation suite of diverse tabular tasks, using TabPFN-3 \citep{grinsztajn2026tabpfn3technicalreport} and TabFM \citep{tabfm2026}. Across both models and all evaluated context budgets, the aligned student outperforms the unaligned baseline in 81\% of cases, with statistically significant gains. With only 10\% of the training context, the aligned models achieve predictive parity with unaligned students provided two to four times as much data. Because activation alignment modifies internal representations rather than inputs or architecture, it is orthogonal to existing compression and context-selection methods; we discuss how the two approaches could be composed in Section~\ref{sec:limitations_and_future_work}.\footnote{Code to reproduce all experiments is available at \url{https://github.com/yoel-zeldes/tabalign}.}

\section{Related Work}
\label{sec:related_work}

\subsection{Tabular Foundation Models and In-Context Learning}

Tabular foundation models perform prediction via in-context learning, processing labeled training examples and test queries in a single forward pass \citep{hollmann2023tabpfn, grinsztajn2026tabpfn3technicalreport}. TabPFN is a Prior-Data Fitted Network \citep{muller2022transformers} trained on synthetic data from structural causal models. While TabPFN~v2 employed feature$\times$item tokenization with alternating attention \citep{hollmann2025accurate}, TabPFN-3 returned to processing tabular data with a single unified token per example, operating across 24 ICL blocks with multi-estimator ensembling. TabFM similarly adopts row-level tokenization with 24 ICL transformer blocks. While tree-based methods such as gradient-boosted decision trees have long outperformed standard deep learning on tabular data \citep{grinsztajn2022tree}, tabular foundation models have emerged that frequently rival or surpass them \citep{mcelfresh2023neural}. Nonetheless, model performance depends heavily on the number of context examples \citep{hollmann2025accurate, qu2025tabicl, ma2024tabdpt}.

\subsection{Compressing Tabular Foundation Models}

Because the computational cost and memory footprint of tabular foundation model inference scale with context size, several lines of research explore model and context compression. On the context dimension, TACO \citep{zabergja2025taco} compresses in-context training examples into compact latent vectors. In a related vein, TuneTables \citep{feuer2024tunetables} optimizes the context itself: it learns, via gradient descent through the frozen backbone, a small set of soft prompt embeddings that distill a large labeled training set into a compact learned context. In contrast to our aligner, these methods modify the model's input rather than its internal states and require GPU backpropagation through the model. Along the architectural dimension, TACTICL \citep{koshil2025tacticl} exploits depth-wise redundancy documented by \citet{gromov2024unreasonable} by pruning transformer layers and substituting them with lightweight adapters, while MotherNet \citep{muller2023mothernet} trains hypernetworks to synthesize small feed-forward networks directly. A complementary strategy acts on context \emph{selection} rather than compression: LoCalPFN \citep{thomas2024localpfn} retrieves the $k$ nearest neighbors of each query from the training set to serve as a per-query local context, and fine-tunes TabPFN end-to-end with those retrieved neighbors in context. 

Our approach shares with TACO the objective of operating over reduced token counts at inference time while retaining the full data's predictive quality. However, existing methods demand computationally intensive offline procedures, such as joint end-to-end pretraining of both compressor and predictor networks (TACO), training layer-replacement adapters with end-to-end fine-tuning (TACTICL), or expensive hypernetwork optimization (MotherNet). In contrast, our linear aligner is trained directly on pre-extracted representations without backpropagating through the model, converging within seconds to minutes on ordinary CPU cores.

\subsection{Feature-Based Knowledge Distillation}

To transfer knowledge from compute- or data-rich regimes without retraining the underlying model, knowledge distillation offers a well-established framework. \citet{hinton2015distilling} introduced matching output-level soft probabilities. Feature-based methods extend this to intermediate representations: FitNets \citep{romero2015fitnets} trains the student to match teacher hidden activations via a learned regressor; attention transfer \citep{zagoruyko2017attention} aligns attention maps; and contrastive representation distillation \citep{tian2020contrastive} uses contrastive objectives for structural knowledge transfer.

These methods distill knowledge across architectures, from a large teacher to a smaller student. In contrast, our work operates within the same frozen architecture, transferring information across data regimes (full context to reduced context) using synthetic unlabeled queries. A further distinction is procedural: in feature-based distillation the learned regressor is training-time scaffolding, discarded once the student's weights are updated, whereas our aligner leaves all weights frozen and is itself the deployed inference-time intervention.

\subsection{Activation Steering}

While distillation provides the objective, the feasibility of modifying intermediate states at test time without retraining is supported by recent advances in representation engineering and activation steering. A growing body of literature demonstrates that hidden states across transformer layers can be directly manipulated during inference to guide downstream outputs. Representation Engineering \citep{zou2023representation} and related activation steering methods demonstrate that high-level concepts are encoded in population-level activations, enabling techniques like Activation Addition \citep{turner2023steering} and Inference-Time Intervention \citep{li2023inference} to shift hidden states along desired directions, such as task intent or truthfulness. Similarly, \citet{arditi2024refusal} showed that refusal mechanisms in language models are mediated by a single linear direction in the residual stream, while function vectors \citep{todd2024function} revealed that entire in-context learning tasks can be compactly localized within intermediate attention head outputs.

Tuned Lens \citep{belrose2023tuned} is particularly related to our approach: it trains an affine probe at each layer of a frozen pretrained model to decode intermediate hidden states into the model's final output distribution. Like our aligner, Tuned Lens learns a linear transformation of a frozen model's intermediate activations. The key difference is in purpose: Tuned Lens is an interpretability tool that reads predictions from intermediate representations, while our aligner writes corrected representations back into the forward pass to improve downstream predictions.

\section{Method}
\label{sec:methods}

\subsection{Problem Formulation}

Let $\mathcal{M}$ be a tabular foundation model that performs in-context learning: given a labeled context set $\mathcal{C} = \{(x_i, y_i)\}_{i=1}^{N}$ and a query $x_q$, the model produces a prediction $\hat{y}_q = \mathcal{M}(x_q \mid \mathcal{C})$ in a single forward pass without parameter updates. We consider the setting where a \emph{teacher} has access to the full context $\mathcal{C}_{\text{full}}$ of size $N_{\text{full}}$, while a \emph{student} is restricted to a subset $\mathcal{C}_{\text{student}} \subset \mathcal{C}_{\text{full}}$ of size $N_{\text{student}} = \lfloor \alpha \cdot N_{\text{full}} \rfloor$, where $\alpha \in (0, 1)$ is the student fraction. The student context $\mathcal{C}_{\text{student}}$ is a stratified random subset of $\mathcal{C}_{\text{full}}$, preserving the label distribution.

Both teacher and student use the same frozen model $\mathcal{M}$. The teacher and student differ only in their context: the student's intermediate representations carry less task information because they are conditioned on fewer labeled examples. We aim to learn a lightweight transformation that maps the student's intermediate activations toward the teacher's, thereby recovering some of the predictive quality lost from the reduced context.

\subsection{Pipeline Overview}

\begin{figure}[t]
\centering
\resizebox{\textwidth}{!}{

\definecolor{tfmteal}{RGB}{31,102,152}
\definecolor{tfmamber}{RGB}{198,120,14}
\definecolor{tfmgreen}{RGB}{26,132,88}
\definecolor{tfmgray}{RGB}{92,97,104}
\definecolor{tfmice}{RGB}{72,142,201}
\definecolor{tfmflame}{RGB}{232,126,34}
\definecolor{tfmflamecore}{RGB}{247,201,72}

\tikzset{
  pics/snow/.style={code={
    \foreach \a in {0,60,120,180,240,300}{
      \draw[tfmice, line width=0.45pt, line cap=round] (0,0) -- (\a:1.80mm);
      \draw[tfmice, line width=0.38pt, line cap=round] (\a:1.20mm) -- ++(\a+50:0.62mm);
      \draw[tfmice, line width=0.38pt, line cap=round] (\a:1.20mm) -- ++(\a-50:0.62mm);
      \draw[tfmice, line width=0.38pt, line cap=round] (\a:0.62mm) -- ++(\a+50:0.42mm);
      \draw[tfmice, line width=0.38pt, line cap=round] (\a:0.62mm) -- ++(\a-50:0.42mm);
    }
  }},
  pics/fire/.style={code={
    \fill[tfmflame] (0,2.05mm)
      .. controls (1.50mm,0.60mm) and (1.35mm,-0.55mm) .. (0,-1.60mm)
      .. controls (-1.35mm,-0.55mm) and (-1.50mm,0.60mm) .. (0,2.05mm);
    \fill[tfmflamecore] (0,0.72mm)
      .. controls (0.78mm,0.05mm) and (0.70mm,-0.62mm) .. (0,-1.46mm)
      .. controls (-0.70mm,-0.62mm) and (-0.78mm,0.05mm) .. (0,0.72mm);
  }},
  pics/dtable/.style n args={3}{code={
    \pgfmathsetmacro{\hL}{0.2*#1}
    \pgfmathsetmacro{\hQ}{0.2*#2}
    \foreach \rowi in {1,...,#1}{
      \draw[fill=#3!9, draw=#3!40, line width=0.25pt]
        (0,-0.2*\rowi+0.2) rectangle (0.95,-0.2*\rowi);
      \fill[#3!70, rounded corners=0.3pt]
        (1.0,-0.2*\rowi+0.175) rectangle (1.2,-0.2*\rowi+0.025);
    }
    \draw[#3!30, line width=0.25pt] (0.317,0)--(0.317,-\hL) (0.633,0)--(0.633,-\hL);
    \foreach \rowi in {1,...,#2}{
      \draw[fill=tfmgray!6, draw=tfmgray!45, line width=0.25pt]
        (0,{-\hL-0.09-0.2*\rowi+0.2}) rectangle (0.95,{-\hL-0.09-0.2*\rowi});
      \pgfmathsetmacro{\ybot}{-\hL-0.09-0.2*\rowi+0.025}
      \foreach \dx in {0, 0.05, 0.10, 0.15, 0.20}{
        \fill[tfmgray!95] ({1.0+\dx},\ybot) circle (0.35pt);
        \fill[tfmgray!95] ({1.0+\dx},{\ybot+0.15}) circle (0.35pt);
      }
      \foreach \dy in {0.05, 0.10}{
        \fill[tfmgray!95] (1.0,{\ybot+\dy}) circle (0.35pt);
        \fill[tfmgray!95] (1.2,{\ybot+\dy}) circle (0.35pt);
      }
    }
    \draw[tfmgray!30, line width=0.25pt]
      (0.317,{-\hL-0.09})--(0.317,{-\hL-0.09-\hQ})
      (0.633,{-\hL-0.09})--(0.633,{-\hL-0.09-\hQ});
    \coordinate (-ctx) at (-0.06,{-0.5*\hL});
    \coordinate (-qry) at (-0.06,{-\hL-0.09-0.5*\hQ});
  }},
  pics/strip/.style n args={1}{code={
    \fill[#1!62] (0,0.5) rectangle (0.3,0.3);
    \fill[#1!20] (0,0.3) rectangle (0.3,0.1);
    \fill[#1!88] (0,0.1) rectangle (0.3,-0.1);
    \fill[#1!38] (0,-0.1) rectangle (0.3,-0.3);
    \fill[#1!13] (0,-0.3) rectangle (0.3,-0.5);
    \draw[#1!75, line width=0.4pt] (0,0.5) rectangle (0.3,-0.5);
  }},
  pics/stackk/.style n args={1}{code={
    \foreach \bx in {0,0.50,1.00}{
      \draw[fill=tfmgray!9, draw=tfmgray!45, line width=0.35pt, rounded corners=0.8pt]
        (\bx,-0.5) rectangle (\bx+0.40,0.5);
    }
    \node[font=\footnotesize, text=tfmgray!85, inner sep=0pt] at (1.68,0) {$\cdots$};
    \draw[fill=#1!16, draw=#1!85, line width=0.7pt, rounded corners=0.8pt]
      (1.96,-0.5) rectangle (2.36,0.5);
  }},
  pics/stackr/.style={code={
    \draw[fill=tfmgray!9, draw=tfmgray!45, line width=0.35pt, rounded corners=0.8pt]
      (0,-0.5) rectangle (0.40,0.5);
    \node[font=\footnotesize, text=tfmgray!85, inner sep=0pt] at (0.675,0) {$\cdots$};
    \foreach \bx in {0.95,1.45}{
      \draw[fill=tfmgray!9, draw=tfmgray!45, line width=0.35pt, rounded corners=0.8pt]
        (\bx,-0.5) rectangle (\bx+0.40,0.5);
    }
  }},
}

\begin{tikzpicture}[
  font=\small,
  >={Stealth[round,length=1.7mm,width=1.4mm]},
  arr/.style={->, line width=0.55pt, draw=tfmgray!85, rounded corners=1.5pt,
              shorten >=2pt, shorten <=2pt},
  tarr/.style={arr, draw=tfmteal!80},
  sarr/.style={arr, draw=tfmamber!85},
  garr/.style={arr, draw=tfmgreen!80},
  dsh/.style={->, line width=0.5pt, draw=tfmgray!65, densely dashed, rounded corners=1.5pt,
              shorten >=2.5pt, shorten <=2.5pt},
  clab/.style={font=\footnotesize, text=tfmgray, inner sep=1pt},
  mlbl/.style={font=\footnotesize, inner sep=1pt},
  algbox/.style={rounded corners=2pt, draw=tfmgreen!80, fill=tfmgreen!10, line width=0.7pt,
                 align=center, inner sep=2.5pt, minimum height=6.2mm},
  lossbox/.style={rounded corners=2pt, draw=tfmgray!60, fill=white, line width=0.5pt,
                  densely dashed, align=center, inner sep=3pt},
  pill/.style={rounded corners=3pt, draw=tfmgray!55, fill=tfmgray!8, line width=0.5pt,
               inner sep=3pt},
  oplus/.style={circle, draw=tfmgreen!85, fill=white, line width=0.65pt, inner sep=0.5pt,
                minimum size=3.6mm, font=\scriptsize},
  ttl/.style={font=\normalsize\bfseries, anchor=south west, inner sep=0pt},
]

\def\yT{1.50}
\def\yS{-0.55}

\pic (tt) at (0,{\yT+0.845}) {dtable={6}{2}{tfmteal}};
\node[clab, anchor=east] at (tt-ctx) {$\mathcal{C}_{\text{full}}$};
\node[clab, anchor=east] at (tt-qry) {$X_{\text{syn}}$};

\draw[tarr] (1.30,\yT) -- (2.08,\yT);
\pic (ts) at (2.15,\yT) {stackk={tfmteal}};
\node[mlbl, text=tfmgray] (mT) at (3.33,{\yT+0.74}) {$\mathcal{M}$};
\pic at ([xshift=-2.7mm]mT.west) {snow};
\node[clab] at (4.31,{\yT-0.74}) {layer $k$};

\draw[tarr] (4.58,\yT) -- (5.30,\yT);
\pic (ta) at (5.35,\yT) {strip={tfmteal}};
\node[mlbl, text=tfmteal, anchor=south] at (5.50,{\yT+0.58}) {$h^{(k)}_{\text{teacher}}$};

\pic (st) at (0,{\yS+0.445}) {dtable={2}{2}{tfmamber}};
\node[clab, anchor=east] at (st-ctx) {$\mathcal{C}_{\text{student}}$};
\node[clab, anchor=east] at (st-qry) {$X_{\text{syn}}$};

\draw[sarr] (1.30,\yS) -- (2.08,\yS);
\pic (ss) at (2.15,\yS) {stackk={tfmamber}};
\node[mlbl, text=tfmgray] (mS) at (3.33,{\yS+0.74}) {$\mathcal{M}$};
\pic at ([xshift=-2.7mm]mS.west) {snow};
\node[clab] at (4.31,{\yS-0.74}) {layer $k$};

\draw[sarr] (4.58,\yS) -- (5.30,\yS);
\pic (sa) at (5.35,\yS) {strip={tfmamber}};
\node[mlbl, text=tfmamber!90!black, anchor=south] at (5.50,{\yS+0.58}) {$h^{(k)}_{\text{student}}$};

\draw[sarr] (5.65,\yS) -- (6.56,\yS);
\fill[tfmamber!85] (6.15,\yS) circle (0.045);
\node[algbox, anchor=west] (falign) at (6.60,\yS)
  {$h \mapsto Wh + b$};
\node[mlbl, text=tfmgreen!70!black, anchor=south] (algA) at ([xshift=1.5mm]falign.north) {aligner $f_\theta$};
\pic at ([xshift=-2.8mm]algA.west) {fire};

\draw[garr] (falign.east) -- (9.22,\yS);
\node[oplus] (plus) at (9.40,\yS) {$+$};
\draw[sarr, shorten <=0pt] (6.15,\yS) -- (6.15,{\yS-0.80}) -- (9.40,{\yS-0.80}) -- (9.40,{\yS-0.20});

\draw[garr] (9.58,\yS) -- (9.95,\yS);
\pic (ha) at (10.00,\yS) {strip={tfmgreen}};
\node[mlbl, text=tfmgreen!70!black, anchor=south] at (10.15,{\yS+0.58}) {$\hat{h}^{(k)}$};

\node[lossbox] (loss) at (12.10,{0.5*(\yT+\yS)}) {$\mathcal{L}_{\text{MSE}}$};
\draw[dsh] (5.65,\yT) -- (12.10,\yT) -- (loss.north);
\draw[dsh] (10.30,\yS) -- (12.10,\yS) -- (loss.south);
\node[clab, anchor=south, fill=tfmgray!4, inner sep=1.5pt] at (8.55,{\yT+0.03}) {target};

\begin{scope}[yshift=-3.20cm]
  \pic (bt) at (0,0.345) {dtable={2}{1}{tfmamber}};
  \node[clab, anchor=east] at (bt-ctx) {$\mathcal{C}_{\text{student}}$};
  \node[clab, anchor=east] at (bt-qry) {$x_q$};

  \draw[sarr] (1.30,0) -- (2.08,0);
  \pic (bs) at (2.15,0) {stackk={tfmamber}};
  \node[mlbl, text=tfmgray] (mB1) at (3.33,0.74) {$\mathcal{M}$};
  \node[clab] at (3.33,-0.74) {layers $1\dots k$};

  \draw[sarr] (4.58,0) -- (5.30,0);
  \pic (bh) at (5.35,0) {strip={tfmamber}};
  \node[mlbl, text=tfmamber!90!black, anchor=south] at (5.50,0.58) {$h^{(k)}$};

  \draw[sarr] (5.65,0) -- (6.56,0);
  \fill[tfmamber!85] (6.15,0) circle (0.045);
  \node[algbox, anchor=west] (bfa) at (6.60,0)
    {$h \mapsto Wh + b$};
  \node[mlbl, text=tfmgreen!70!black, anchor=south] at (bfa.north) {aligner $f_\theta$};

  \draw[garr] (bfa.east) -- (9.22,0);
  \node[oplus] (bplus) at (9.40,0) {$+$};
  \draw[sarr, shorten <=0pt] (6.15,0) -- (6.15,-0.80) -- (9.40,-0.80) -- (9.40,-0.20);

  \draw[garr] (9.58,0) -- (9.95,0);
  \pic (bhh) at (10.00,0) {strip={tfmgreen}};
  \node[mlbl, text=tfmgreen!70!black, anchor=south] at (10.15,0.58) {$\hat{h}^{(k)}$};

  \draw[garr] (10.30,0) -- (10.88,0);
  \pic (br) at (10.93,0) {stackr};
  \node[mlbl, text=tfmgray] (mB2) at (11.86,0.74) {$\mathcal{M}$};
  \node[clab] at (11.899,-0.74) {layers $k{+}1\dots$};

  \draw[arr] (12.86,0) -- (13.32,0);
  \node[pill] (yh) at (13.67,0) {$\hat{y}_q$};
\end{scope}

\begin{scope}[on background layer]
  \fill[tfmgray!4, rounded corners=3pt] (-1.15,-1.50) rectangle (14.10,2.63);
  \fill[tfmgray!4, rounded corners=3pt] (-1.15,-4.20) rectangle (14.10,-2.22);
\end{scope}
\node[ttl] at (-1.15,2.72) {(a)\; Offline: fit the aligner};
\node[ttl] at (-1.15,-2.13) {(b)\; Inference: correct the activation};

\end{tikzpicture}}
\caption{\textbf{(a)}~Unlabeled synthetic queries $X_{\text{syn}}$ are appended to the full context (teacher) and to the reduced context (student), and both are passed through the same frozen model $\mathcal{M}$. A linear aligner $f_\theta$ is fit by MSE so that the corrected student activation matches the teacher's at layer $k$. \textbf{(b)}~At test time, the trained aligner is applied to the student's layer-$k$ activations.}
\label{fig:method_overview}
\end{figure}

The method consists of four stages (Figure~\ref{fig:method_overview}):

\begin{enumerate}
    \item \textbf{Synthetic data generation}: Generate unlabeled query vectors $X_{\text{syn}}$ from the feature distribution, requiring no additional labels.
    \item \textbf{Activation extraction}: Pass $X_{\text{syn}}$ through $\mathcal{M}$ under both teacher and student contexts, extracting intermediate activations at a chosen layer $k$.
    \item \textbf{Aligner training}: Fit a linear model to map student activations to teacher activations using MSE loss on the extracted activation pairs.
    \item \textbf{Inference-time alignment}: Apply the trained aligner to transform the student's activations at layer $k$ during the forward pass.
\end{enumerate}

\subsection{Synthetic Data Generation}

To train the aligner without requiring additional labeled data, we generate $N_{\text{syn}}$ synthetic query vectors $X_{\text{syn}} \in \mathbb{R}^{N_{\text{syn}} \times m}$ (where $m$ is the number of features) by using TabPFN's unsupervised generative capabilities to model the training features $X_{\text{train}}$ and sampling from the learned joint multivariate distribution. This captures feature correlations present in the real data. In our experiments, we use $N_{\text{syn}} = 1{,}000$ synthetic samples. For TabFM, we use the same TabPFN-generated synthetic queries; as the aligner operates on model-internal activations, the synthetic input distribution serves only to probe the representation space and is not model-specific.

\subsection{Aligner Training}

The aligner $f_\theta$ is a linear model $f_\theta(h) = Wh + b$, where $W \in \mathbb{R}^{d \times d}$ and $b \in \mathbb{R}^d$, operating on intermediate representations of hidden dimension $d$ at a designated layer $k$.

\paragraph{Residual formulation.} Rather than predicting the teacher activation directly, the aligner predicts the residual $\Delta h_i = h_{\text{teacher},i}^{(k)} - h_{\text{student},i}^{(k)}$, where $h_{\text{teacher},i}^{(k)}, h_{\text{student},i}^{(k)} \in \mathbb{R}^{d}$ denote the layer-$k$ activations of the $i$-th synthetic query under the full context $\mathcal{C}_{\text{full}}$ and the student context $\mathcal{C}_{\text{student}}$, respectively, yielding $N_{\text{syn}}$ activation pairs. The weights are initialized with near-zero values, so that the aligner initially outputs near-zero corrections, preserving the student's original representations at the start of training. At inference, the aligned activation is $\hat{h}^{(k)} = h_{\text{student}}^{(k)} + f_\theta(h_{\text{student}}^{(k)})$.

\paragraph{Layer selection.} For both TabPFN and TabFM (each comprising 24 layers), we extract and align activations at the final layer.

\paragraph{Training objective.} The aligner is trained to minimize MSE loss:
\begin{equation}
\mathcal{L} = \frac{1}{N_{\text{syn}}} \sum_{i=1}^{N_{\text{syn}}} \left\| f_\theta\!\left(h_{\text{student},i}^{(k)}\right) - \Delta h_i \right\|_2^2
\label{eq:mse_loss}
\end{equation}

\paragraph{Optimization.} We use AdamW with a learning rate of $10^{-3}$ for TabPFN and $10^{-4}$ for TabFM, and early stopping with a patience of 10 epochs on a held-out validation split (80/20 random split of the activation pairs).

\paragraph{Multi-estimator models.} We use an 8-estimator ensemble for TabPFN and a single estimator for TabFM. For TabPFN, a separate aligner is trained independently for each estimator.

\paragraph{Shared preprocessing.} Tabular foundation models rely on data-dependent preprocessing pipelines, such as quantile transforms, categorical encodings, and randomized feature subsets or permutations per ensemble member. To ensure consistent representations, the student inherits the teacher's fitted preprocessors while conditioning its in-context learning strictly on the reduced subset $\mathcal{C}_{\text{student}}$. This guarantees that each student estimator operates in the same input representation space as its teacher counterpart, avoiding representation drift caused by differing random permutations, mismatched category codes, or unstable quantile estimates on small sample subsets, which would introduce arbitrary distortions and render the alignment task ineffective.

\section{Experimental Setup}
\label{sec:eval_protocol}

\paragraph{Benchmark.} We evaluate on 38 binary and multiclass classification datasets from the TabArena-v0.1 benchmark suite \citep{erickson2025tabarena}.

\paragraph{Student fractions.} For each dataset, we vary the student fraction $\alpha \in \{0.1, 0.2, 0.3, 0.4, 0.5\}$, where $N_{\text{student}} = \lfloor \alpha \cdot N_{\text{full}} \rfloor$. We focus our primary analysis on context reductions of $2\times$ to $10\times$ ($\alpha \le 0.5$), where context truncation delivers the most critical computational and memory savings. For completeness, evaluations across the full spectrum $\alpha \in \{0.1, \dots, 0.9\}$ are provided in Table~\ref{tab:combined_per_dataset}, with consistent conclusions.
 
 \paragraph{Repeats.} Each experiment is repeated 5 times using the predefined train/test splits attached to each dataset's OpenML task \citep{vanschoren2014openml}, to account for variability.
 
 \paragraph{Metric.} We use ROC-AUC for binary classification datasets and negative log-loss for multiclass datasets (higher is better).
 
 \paragraph{Baselines.} We compare four conditions at each student fraction:
 \begin{enumerate}
     \item \textbf{Baseline student}: $\mathcal{M}$ with context $\mathcal{C}_{\text{student}}$ (no alignment).
     \item \textbf{Aligned student}: $\mathcal{M}$ with context $\mathcal{C}_{\text{student}}$ and the trained aligner applied at the last layer.
     \item \textbf{Teacher}: $\mathcal{M}$ with the full context $\mathcal{C}_{\text{full}}$ (upper bound).
     \item \textbf{XGBoost}: XGBoost \citep{chen2016xgboost} trained on $\mathcal{C}_{\text{full}}$ using the standard default hyperparameters established in \citet{gorishniy2021revisiting}. This represents the standard practical alternative when full-context transformer inference is constrained by compute or memory, as tree-based models can utilize the entire dataset without incurring higher inference latency.
 \end{enumerate}
 
 \paragraph{Effective sample fraction.} To quantify the alignment benefit, we compute an effective sample fraction $E_\alpha$: the student fraction at which the unaligned student would need to operate to achieve the same metric score as the aligned student at fraction $\alpha$. This is obtained by linear interpolation between baseline scores at neighboring fractions. An aligned student with $E_\alpha > \alpha$ is effectively operating as if it had more data.
 
 \paragraph{Models tested.} We evaluate on two tabular foundation models: TabPFN-3 with an 8-estimator ensemble and TabFM with a single estimator.
 
 \paragraph{Teacher superiority condition.} We evaluate alignment metrics on instances where the teacher outperforms the unaligned baseline. When a student already matches or surpasses the full-context teacher (e.g., due to split variance on small datasets), distilling teacher representations is unmotivated. Across all $38 \times 5 = 190$ evaluation slices (38 datasets across 5 context fractions), this condition excludes only 5 slices for TabPFN and 2 for TabFM.

\section{Results}
\label{sec:results}

We evaluate activation alignment across three central empirical questions: (1) Does aligning intermediate representations systematically improve downstream performance over the unaligned baseline? (2) How does the effective data gain behave across varying context constraints? (3) Can a data-constrained foundation model with alignment compete with or surpass a strong tree-based model trained on the entire dataset?

\subsection{Alignment Substantially Improves Over Baseline Student}

\begin{figure}[t]
 \centering
 \includegraphics[width=\textwidth]{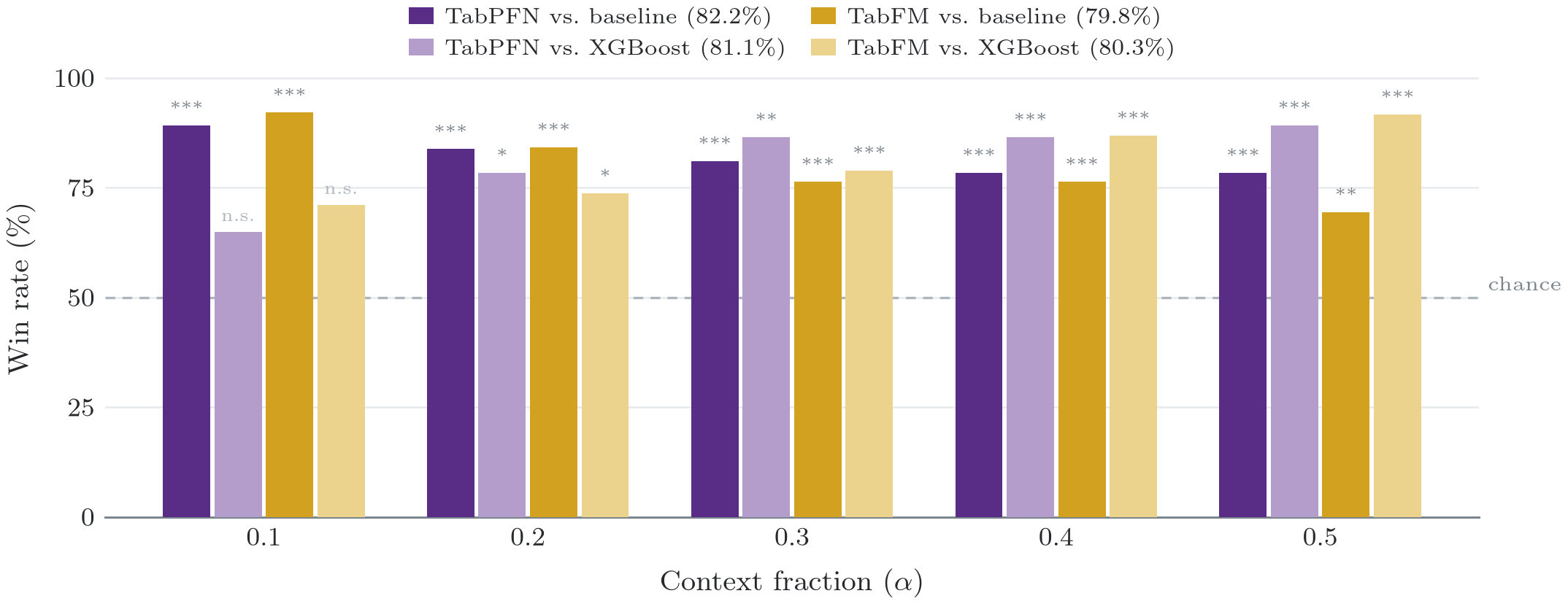}
 \caption{Pairwise win rates of the aligned student over the unaligned baseline student and full-training-budget XGBoost across context fractions $\alpha \in \{0.1, \dots, 0.5\}$ on 38 TabArena classification datasets (5 repeats) for TabPFN and TabFM. Overall win rates across all context budgets are reported in the legend (82.2\% vs.\ base and 81.1\% vs.\ XGBoost for TabPFN; 79.8\% vs.\ base and 80.3\% vs.\ XGBoost for TabFM). Statistical significance annotations above each bar indicate two-sided paired Wilcoxon signed-rank test results against the respective comparator ($^{***} p < 0.001$, $^{**} p < 0.01$, $^{*} p < 0.05$; \textrm{n.s.}: not significant).}
 \label{fig:combined_win_rate}
 \end{figure}

\begin{figure}[t]
\centering
\includegraphics[width=0.75\textwidth]{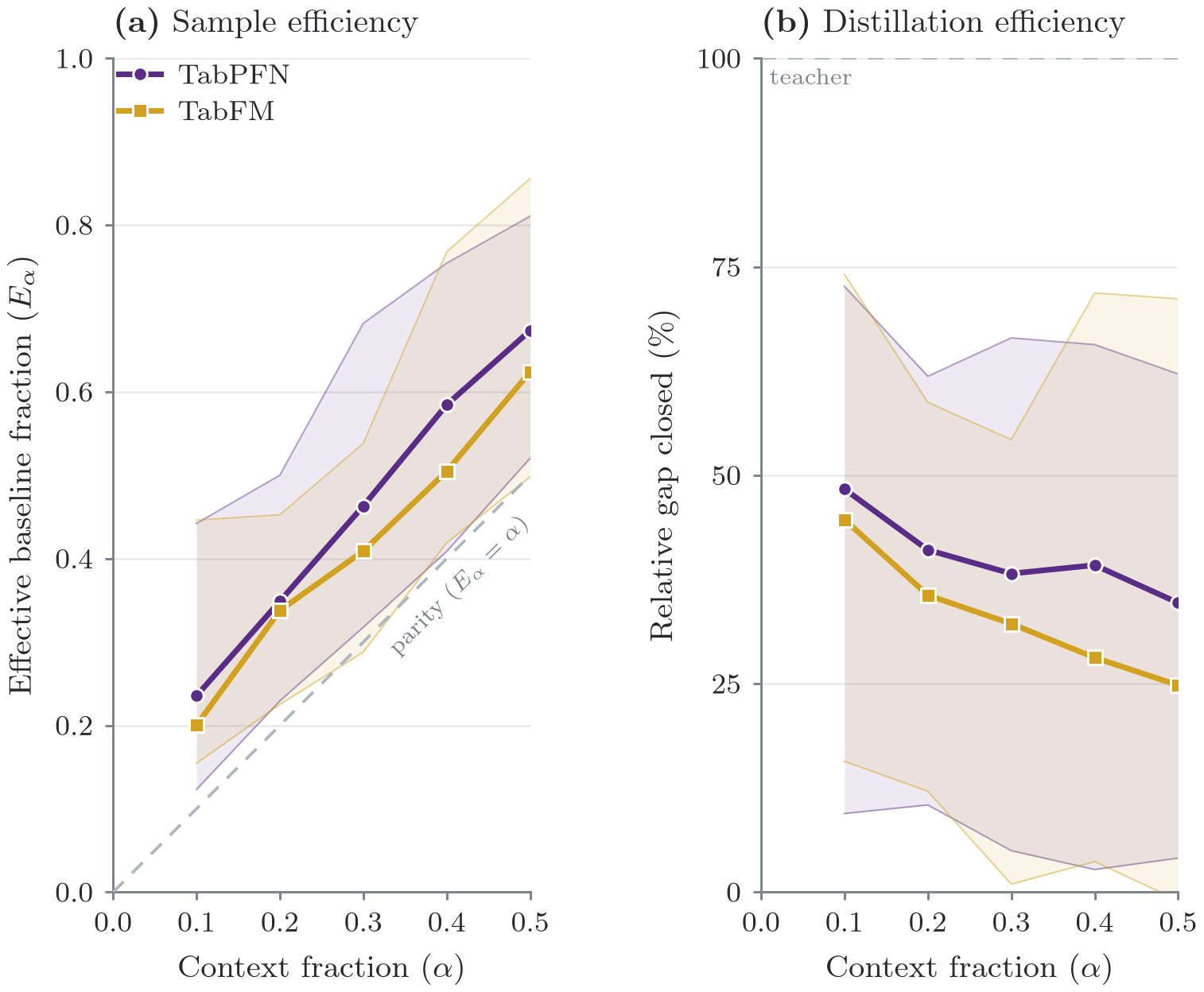}
\caption{Median scaling and distillation efficiency across 38 TabArena classification datasets for TabPFN and TabFM, averaged over 5 repeats. Shaded bands denote the 25th--75th percentiles across datasets. \textbf{(a) Sample Efficiency}: Median effective baseline context fraction $E_\alpha$ achieved by the aligned student as a function of the actual context budget $\alpha$. Both models achieve positive data gains across all context fractions. \textbf{(b) Distillation Efficiency}: Median relative gap closed between baseline student and full-context teacher across context fractions $\alpha$, showing that alignment closes a significant fraction of the gap across all context budgets.}
\label{fig:combined_scaling}
\end{figure}

Figure~\ref{fig:combined_win_rate} presents pairwise win rates with Wilcoxon signed-rank tests against both the unaligned baseline student and full-data XGBoost, and Figure~\ref{fig:combined_scaling} illustrates the scaling curves and distillation dynamics (detailed per-dataset benchmarks across all context budgets $\alpha \in \{0.1, \dots, 0.9\}$ are reported in Table~\ref{tab:combined_per_dataset}). 

Across diverse tabular tasks, intermediate activation alignment yields broad, statistically significant performance gains over the unaligned baseline student (Figure~\ref{fig:combined_win_rate}). Statistical significance is evaluated via a two-sided paired Wilcoxon signed-rank test across datasets \citep{demsar2006statistical}: for each individual context fraction $\alpha$, the test pairs the 5-repeat mean scores of aligned models against the respective comparator (the unaligned baseline student or full-context XGBoost) across all evaluated datasets. To evaluate overall statistical significance across all context budgets without pseudoreplication (i.e., violating the independence assumption by treating multiple context fractions of the same dataset as independent observations), the test pairs the per-dataset mean scores averaged across all context fractions. Against the unaligned baseline student, TabPFN achieves an 82.2\% win rate over all evaluated (dataset, context fraction) pairs, with performance improvements being statistically significant overall ($p < 0.0001$) and across every individual context fraction. For TabFM, the aligned student achieves a 79.8\% overall win rate against the baseline student ($p < 0.0001$), maintaining statistically significant gains across every evaluated context fraction. 
 
\subsection{Sample Efficiency and Distillation Gains}

To understand how the aligner modulates task representations under varying data constraints, we examine the effective sample fraction $E_\alpha$ and the proportion of the teacher--student gap recovered (Figure~\ref{fig:combined_scaling}).

\paragraph{Data amplification in the low-data regime.} In extreme low-data settings ($\alpha = 0.1$, corresponding to 10\% of available training points), the aligned student achieves a median effective sample fraction of $E_{0.1} = 0.235$ for TabPFN and $0.200$ for TabFM (with mean $E_{0.1} \approx 0.33$ across both architectures, as can be seen in Table~\ref{tab:combined_per_dataset}). Without processing additional in-context tokens during inference, the aligned student attains predictive accuracy comparable to an unaligned model provided with two to four times as many labeled examples. Crucially, median sample efficiency remains strictly above parity across all evaluated context budgets.

\paragraph{Gap closure across context budgets.} Examining the relative gap closed reveals consistent recovery of the teacher's predictive advantage: across all evaluated context fractions, the median gap closed remains above 34\% for TabPFN and above 24\% for TabFM. While alignment provides non-trivial benefits across the entire context spectrum, it is particularly effective in lower-data regimes: at $\alpha = 0.1$, alignment recovers a median of 48.3\% and 44.7\% of the teacher--student performance deficit for TabPFN and TabFM, respectively.

\subsection{Comparison with Full-Data XGBoost}

A central practical question for tabular foundation models is whether context-constrained in-context learning can match standard tree-based models trained on the complete dataset. While XGBoost had access to 100\% of the training data ($\mathcal{C}_{\text{full}}$), the aligned student was restricted to subsets between 10\% and 50\% of the context examples.

\begin{figure}[t]
\centering
\includegraphics[width=\textwidth]{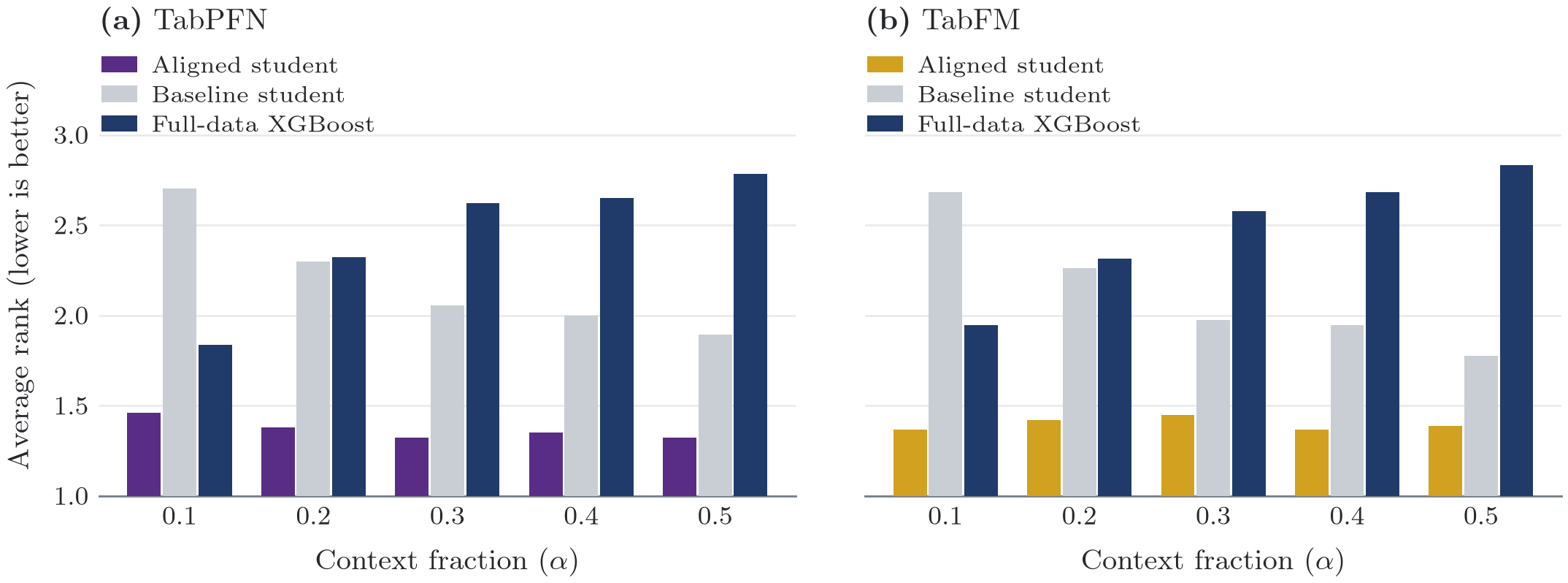}
\caption{3-way average rank (lower is better) across 38 TabArena classification datasets for (a) TabPFN and (b) TabFM, averaged over 5 repeats. At each context fraction $\alpha \in \{0.1, \dots, 0.5\}$, bars compare the aligned student, unaligned baseline student, and full-context XGBoost. Across all context budgets, the aligned student maintains the top average ranking.}
\label{fig:avg_rank_histogram}
\end{figure}

Despite this substantial data handicap, the aligned student broadly outperforms full-training-budget XGBoost across both architectures (Figure~\ref{fig:combined_win_rate}). Overall, the aligned student beats full-data XGBoost in 81.1\% of conditions for TabPFN and 80.3\% for TabFM ($p < 0.0001$ overall for both models), maintaining statistically significant advantages across all context budgets $\alpha \ge 0.2$. In a 3-way ranking (aligned student vs.\ baseline student vs.\ full-data XGBoost), the aligned student maintains the top average rank across all context budgets: 1.37 for TabPFN and 1.40 for TabFM, compared to 2.19 and 2.13 for the baseline student and 2.44 and 2.47 for XGBoost (Figure~\ref{fig:avg_rank_histogram}). These results demonstrate that combining tabular ICL models with activation alignment can surpass fully trained gradient boosting models even under strict context budgets.

\section{Limitations and Future Work}
\label{sec:limitations_and_future_work}

\subsection{Limitations}

Several limitations bound the scope of these findings:

\begin{enumerate}
    \item \textbf{Classification only.} While our evaluation focused on classification tasks, the aligner operates entirely at the level of intermediate transformer activations and uses unlabeled synthetic queries, with no classification-specific assumptions. We therefore expect the method to translate directly to regression settings, though this has yet to be empirically verified.

    \item \textbf{Model coverage.} Our empirical evaluation was conducted on two representative tabular foundation models (TabPFN-3 and TabFM). We expect that activation alignment generalizes to other tabular in-context learning architectures, but validating this across a broader family of foundation models remains an important empirical step.

    \item \textbf{Not universally beneficial.} Five of the 38 datasets (MIC, SDSS17, credit\_card\_clients\_default, kddcup09\_appetency, and polish\_companies\_bankruptcy) show a negative mean effective-sample gain for both models; Table~\ref{tab:combined_per_dataset} provides the full per-dataset breakdown. We do not currently have a characterization that separates these datasets from those that benefit. Identifying such properties remains an open question and would allow practitioners to determine in advance whether alignment is worth applying to a given task.
\end{enumerate}

\subsection{Future Directions}

The results suggest several directions for future work:

\begin{itemize}
    \item \textbf{Expressive aligners and training objectives.} Future work could explore non-linear aligners (e.g., multi-layer perceptrons), adaptive layer selection, intervening across multiple layers simultaneously, systematic hyperparameter optimization, and alternative training objectives such as incorporating an end-to-end distillation loss over output logits.
    \item \textbf{Combination with compression and context selection.} Activation alignment could be applied to models compressed via context compression (e.g., TACO) or layer pruning (e.g., TACTICL). Because compression inevitably perturbs intermediate activations away from those of the uncompressed model, learning an aligner to map the compressed model's representations back toward the full-capacity teacher could recover performance lost to compression while retaining inference speedups. The same composition applies to retrieval-based context selection \citep[e.g.,][]{thomas2024localpfn}: our student's context is a stratified random subset and we do not compare against selection strategies, but an aligner could equally correct the representations of a student whose context was chosen by retrieval.
    \item \textbf{Mechanistic analysis.} While our results establish that a linear correction is sufficient to recover a substantial share of the teacher--student gap, they do not characterize the geometric structure of the representation displacement. Future mechanistic work analyzing intermediate representation spaces across architectures could clarify why linear alignment is effective, whether non-linear structure remains uncaptured, and why different models exhibit differing degrees of alignment gains.
    \item \textbf{Aligners for genuinely small datasets.} Currently, the aligner requires offline access to the full dataset to extract teacher activations, which addresses inference-time compute and memory bottlenecks but does not apply when only a small dataset exists to begin with. The framework could be extended to genuinely low-data regimes by conditioning the aligner on summary statistics or domain priors (e.g., population moments that differ from small-sample estimates). An aligner could be trained across diverse external datasets where abundant data is available, learning to map student activations and summary statistics to teacher representations. At test time, this model could then align representations on genuinely small, unseen datasets where additional examples do not exist.
\end{itemize}

\section{Conclusion}
\label{sec:conclusion}

In this work, we introduce activation alignment to address the sample efficiency bottleneck in tabular in-context learning. By projecting intermediate hidden states of data-limited student models toward those elicited under full context at a single layer, our approach enables substantially higher predictive performance from small context budgets. The aligner is parameterized as an affine transformation optimized via MSE over synthetic data, requiring no external labels, running on commodity hardware, and leaving the underlying backbone frozen.

Across 38 classification benchmarks from TabArena evaluated on both TabPFN-3 and TabFM, activation alignment yields broad, statistically significant gains over unaligned baselines. Gains are particularly striking in low-data regimes: with only 10\% of training examples as context, aligned students match the predictive accuracy of unaligned models provided two to four times as much data. Furthermore, aligned students operating on partial contexts systematically outperform full-data gradient-boosted trees in over 80\% of evaluations.

More broadly, our findings demonstrate that intermediate representations in tabular foundation models are readily steerable toward high-data regimes using simple linear mappings. Activation alignment offers a practical, model-agnostic technique for overcoming memory and context limitations in real-world tabular machine learning.

\section*{Acknowledgments}
The author thanks Alan Arazi (Technion -- Israel Institute of Technology) for valuable discussions and feedback.

\bibliography{references}
\bibliographystyle{tmlr}

\clearpage
\appendix
\section{Per-Dataset Sample Efficiency Breakdown}
\label{app:per_dataset}

Table~\ref{tab:combined_per_dataset} reports the detailed per-dataset effective baseline sample fraction $E_\alpha$ across the full context spectrum $\alpha \in \{0.1, \dots, 0.9\}$ for all 38 TabArena classification datasets.

\begin{landscape}
\begin{table}[p]
\caption{Per-dataset effective baseline sample fraction $E_\alpha$ and Mean Gain ($E_\alpha - \alpha$) across 38 TabArena classification datasets for TabPFN (PFN) and TabFM (FM). Shaded green cells indicate effective samples ($E_\alpha > \alpha$ or Mean Gain $> 0$). Dashes (---) denote slices excluded by the teacher superiority filter (Section~\ref{sec:eval_protocol}).}
\label{tab:combined_per_dataset}
\begin{center}
\scriptsize
\setlength{\tabcolsep}{2.5pt}
\renewcommand{\arraystretch}{0.85}
\begin{tabular}{@{}lcccccccccccccccccccc@{}}
\toprule
Dataset & \multicolumn{2}{c}{$0.1$} & \multicolumn{2}{c}{$0.2$} & \multicolumn{2}{c}{$0.3$} & \multicolumn{2}{c}{$0.4$} & \multicolumn{2}{c}{$0.5$} & \multicolumn{2}{c}{$0.6$} & \multicolumn{2}{c}{$0.7$} & \multicolumn{2}{c}{$0.8$} & \multicolumn{2}{c}{$0.9$} & \multicolumn{2}{c}{Mean Gain} \\
 & \textbf{PFN} & \textbf{FM} & \textbf{PFN} & \textbf{FM} & \textbf{PFN} & \textbf{FM} & \textbf{PFN} & \textbf{FM} & \textbf{PFN} & \textbf{FM} & \textbf{PFN} & \textbf{FM} & \textbf{PFN} & \textbf{FM} & \textbf{PFN} & \textbf{FM} & \textbf{PFN} & \textbf{FM} & \textbf{PFN} & \textbf{FM} \\
\midrule
APSFailure & 0.05 & \cellcolor{lightgreen}0.17 & \cellcolor{lightgreen}0.22 & \cellcolor{lightgreen}0.24 & 0.28 & 0.27 & 0.38 & \cellcolor{lightgreen}0.65 & 0.50 & 0.25 & 0.59 & \cellcolor{lightgreen}0.70 & 0.70 & \cellcolor{lightgreen}0.70 & 0.69 & 0.72 & \cellcolor{lightgreen}0.91 & --- & -0.02 & \cellcolor{lightgreen}+0.01 \\
Amazon\_employee\_access & \cellcolor{lightgreen}0.12 & \cellcolor{lightgreen}0.13 & 0.20 & \cellcolor{lightgreen}0.21 & \cellcolor{lightgreen}0.30 & \cellcolor{lightgreen}0.31 & \cellcolor{lightgreen}0.41 & \cellcolor{lightgreen}0.42 & \cellcolor{lightgreen}0.52 & \cellcolor{lightgreen}0.53 & \cellcolor{lightgreen}0.62 & \cellcolor{lightgreen}0.62 & \cellcolor{lightgreen}0.72 & \cellcolor{lightgreen}0.72 & \cellcolor{lightgreen}0.83 & \cellcolor{lightgreen}0.83 & \cellcolor{lightgreen}0.93 & \cellcolor{lightgreen}0.93 & \cellcolor{lightgreen}+0.02 & \cellcolor{lightgreen}+0.02 \\
Bank\_Customer\_Churn & \cellcolor{lightgreen}0.29 & \cellcolor{lightgreen}0.20 & \cellcolor{lightgreen}0.40 & \cellcolor{lightgreen}0.29 & \cellcolor{lightgreen}0.58 & \cellcolor{lightgreen}0.45 & \cellcolor{lightgreen}0.68 & \cellcolor{lightgreen}0.61 & \cellcolor{lightgreen}0.74 & \cellcolor{lightgreen}0.91 & \cellcolor{lightgreen}0.80 & \cellcolor{lightgreen}0.84 & \cellcolor{lightgreen}0.85 & \cellcolor{lightgreen}0.94 & \cellcolor{lightgreen}0.89 & \cellcolor{lightgreen}0.94 & \cellcolor{lightgreen}0.92 & \cellcolor{lightgreen}0.95 & \cellcolor{lightgreen}+0.18 & \cellcolor{lightgreen}+0.18 \\
Bioresponse & \cellcolor{lightgreen}0.12 & \cellcolor{lightgreen}0.13 & 0.19 & \cellcolor{lightgreen}0.24 & 0.26 & \cellcolor{lightgreen}0.35 & 0.36 & \cellcolor{lightgreen}0.44 & 0.46 & 0.50 & 0.55 & 0.58 & 0.64 & 0.69 & 0.76 & \cellcolor{lightgreen}0.81 & \cellcolor{lightgreen}0.91 & \cellcolor{lightgreen}0.91 & -0.03 & \cellcolor{lightgreen}+0.02 \\
Diabetes130US & \cellcolor{lightgreen}0.16 & \cellcolor{lightgreen}0.17 & \cellcolor{lightgreen}0.25 & \cellcolor{lightgreen}0.34 & \cellcolor{lightgreen}0.34 & \cellcolor{lightgreen}0.47 & \cellcolor{lightgreen}0.42 & \cellcolor{lightgreen}0.70 & 0.49 & \cellcolor{lightgreen}0.70 & 0.58 & \cellcolor{lightgreen}0.74 & \cellcolor{lightgreen}0.72 & \cellcolor{lightgreen}0.78 & \cellcolor{lightgreen}0.82 & \cellcolor{lightgreen}0.82 & 0.89 & \cellcolor{lightgreen}0.93 & \cellcolor{lightgreen}+0.02 & \cellcolor{lightgreen}+0.13 \\
E-CommereShippingData & \cellcolor{lightgreen}0.91 & 0.10 & \cellcolor{lightgreen}0.46 & 0.10 & \cellcolor{lightgreen}1.00 & 0.10 & \cellcolor{lightgreen}0.98 & 0.10 & \cellcolor{lightgreen}1.00 & --- & \cellcolor{lightgreen}1.00 & --- & \cellcolor{lightgreen}1.00 & \cellcolor{lightgreen}0.85 & \cellcolor{lightgreen}0.96 & \cellcolor{lightgreen}0.91 & \cellcolor{lightgreen}0.94 & --- & \cellcolor{lightgreen}+0.42 & -0.06 \\
Fitness\_Club & \cellcolor{lightgreen}0.90 & \cellcolor{lightgreen}0.90 & \cellcolor{lightgreen}0.90 & \cellcolor{lightgreen}0.90 & \cellcolor{lightgreen}0.90 & \cellcolor{lightgreen}0.90 & \cellcolor{lightgreen}0.90 & \cellcolor{lightgreen}0.90 & \cellcolor{lightgreen}0.90 & \cellcolor{lightgreen}0.90 & --- & --- & \cellcolor{lightgreen}0.90 & --- & \cellcolor{lightgreen}0.90 & --- & 0.90 & 0.90 & \cellcolor{lightgreen}+0.41 & \cellcolor{lightgreen}+0.50 \\
GiveMeSomeCredit & \cellcolor{lightgreen}0.21 & \cellcolor{lightgreen}0.19 & \cellcolor{lightgreen}0.30 & \cellcolor{lightgreen}0.31 & \cellcolor{lightgreen}0.40 & \cellcolor{lightgreen}0.43 & \cellcolor{lightgreen}0.46 & \cellcolor{lightgreen}0.46 & \cellcolor{lightgreen}0.51 & \cellcolor{lightgreen}0.55 & \cellcolor{lightgreen}0.64 & 0.55 & 0.69 & 0.62 & 0.78 & 0.74 & 0.86 & 0.88 & \cellcolor{lightgreen}+0.04 & \cellcolor{lightgreen}+0.03 \\
HR\_Analytics\_Job\_Change\_of\_Data\_Scientists & \cellcolor{lightgreen}0.40 & \cellcolor{lightgreen}0.21 & \cellcolor{lightgreen}0.52 & \cellcolor{lightgreen}0.36 & \cellcolor{lightgreen}0.71 & \cellcolor{lightgreen}0.40 & \cellcolor{lightgreen}0.75 & 0.38 & \cellcolor{lightgreen}0.77 & \cellcolor{lightgreen}0.62 & \cellcolor{lightgreen}0.78 & \cellcolor{lightgreen}0.67 & \cellcolor{lightgreen}0.77 & \cellcolor{lightgreen}0.75 & \cellcolor{lightgreen}0.89 & \cellcolor{lightgreen}0.81 & \cellcolor{lightgreen}0.95 & \cellcolor{lightgreen}0.97 & \cellcolor{lightgreen}+0.23 & \cellcolor{lightgreen}+0.07 \\
Is-this-a-good-customer & \cellcolor{lightgreen}0.47 & \cellcolor{lightgreen}0.95 & \cellcolor{lightgreen}0.89 & \cellcolor{lightgreen}0.93 & \cellcolor{lightgreen}0.97 & \cellcolor{lightgreen}1.00 & \cellcolor{lightgreen}1.00 & \cellcolor{lightgreen}1.00 & \cellcolor{lightgreen}0.98 & \cellcolor{lightgreen}1.00 & \cellcolor{lightgreen}0.99 & \cellcolor{lightgreen}1.00 & \cellcolor{lightgreen}0.98 & \cellcolor{lightgreen}1.00 & \cellcolor{lightgreen}0.97 & \cellcolor{lightgreen}0.99 & \cellcolor{lightgreen}1.00 & \cellcolor{lightgreen}0.96 & \cellcolor{lightgreen}+0.42 & \cellcolor{lightgreen}+0.48 \\
MIC & 0.00 & 0.04 & 0.13 & 0.16 & 0.28 & 0.23 & 0.30 & 0.26 & 0.37 & 0.35 & 0.48 & 0.41 & 0.49 & 0.44 & 0.79 & 0.43 & 0.86 & 0.65 & -0.09 & -0.17 \\
Marketing\_Campaign & \cellcolor{lightgreen}0.69 & \cellcolor{lightgreen}0.69 & \cellcolor{lightgreen}0.79 & \cellcolor{lightgreen}0.62 & \cellcolor{lightgreen}0.80 & \cellcolor{lightgreen}0.84 & \cellcolor{lightgreen}0.73 & \cellcolor{lightgreen}0.80 & \cellcolor{lightgreen}0.68 & \cellcolor{lightgreen}0.80 & \cellcolor{lightgreen}0.78 & \cellcolor{lightgreen}0.80 & \cellcolor{lightgreen}0.84 & \cellcolor{lightgreen}0.81 & \cellcolor{lightgreen}0.88 & \cellcolor{lightgreen}0.85 & \cellcolor{lightgreen}1.00 & \cellcolor{lightgreen}0.92 & \cellcolor{lightgreen}+0.30 & \cellcolor{lightgreen}+0.29 \\
NATICUSdroid & \cellcolor{lightgreen}0.22 & \cellcolor{lightgreen}0.23 & \cellcolor{lightgreen}0.35 & \cellcolor{lightgreen}0.34 & \cellcolor{lightgreen}0.44 & \cellcolor{lightgreen}0.41 & \cellcolor{lightgreen}0.53 & \cellcolor{lightgreen}0.51 & \cellcolor{lightgreen}0.58 & \cellcolor{lightgreen}0.58 & \cellcolor{lightgreen}0.64 & \cellcolor{lightgreen}0.63 & \cellcolor{lightgreen}0.82 & \cellcolor{lightgreen}0.75 & \cellcolor{lightgreen}0.84 & \cellcolor{lightgreen}0.82 & \cellcolor{lightgreen}0.91 & \cellcolor{lightgreen}0.92 & \cellcolor{lightgreen}+0.09 & \cellcolor{lightgreen}+0.08 \\
SDSS17 & 0.06 & \cellcolor{lightgreen}0.13 & 0.16 & \cellcolor{lightgreen}0.20 & 0.23 & 0.28 & 0.30 & 0.39 & 0.42 & 0.48 & \cellcolor{lightgreen}0.60 & 0.56 & 0.69 & \cellcolor{lightgreen}0.71 & 0.80 & 0.79 & 0.85 & \cellcolor{lightgreen}0.91 & -0.04 & -0.00 \\
anneal & \cellcolor{lightgreen}0.12 & \cellcolor{lightgreen}0.18 & \cellcolor{lightgreen}0.27 & \cellcolor{lightgreen}0.35 & \cellcolor{lightgreen}0.44 & \cellcolor{lightgreen}0.41 & \cellcolor{lightgreen}0.53 & \cellcolor{lightgreen}0.52 & \cellcolor{lightgreen}0.58 & \cellcolor{lightgreen}0.58 & \cellcolor{lightgreen}0.74 & \cellcolor{lightgreen}0.77 & \cellcolor{lightgreen}0.84 & \cellcolor{lightgreen}0.84 & \cellcolor{lightgreen}0.90 & \cellcolor{lightgreen}0.91 & \cellcolor{lightgreen}0.96 & \cellcolor{lightgreen}0.96 & \cellcolor{lightgreen}+0.10 & \cellcolor{lightgreen}+0.11 \\
bank-marketing & \cellcolor{lightgreen}0.44 & \cellcolor{lightgreen}0.35 & \cellcolor{lightgreen}0.67 & \cellcolor{lightgreen}0.39 & \cellcolor{lightgreen}0.69 & \cellcolor{lightgreen}0.45 & \cellcolor{lightgreen}0.75 & \cellcolor{lightgreen}0.60 & \cellcolor{lightgreen}0.78 & \cellcolor{lightgreen}0.63 & \cellcolor{lightgreen}0.83 & \cellcolor{lightgreen}0.66 & \cellcolor{lightgreen}0.89 & \cellcolor{lightgreen}0.80 & \cellcolor{lightgreen}0.95 & \cellcolor{lightgreen}0.89 & \cellcolor{lightgreen}0.96 & \cellcolor{lightgreen}0.95 & \cellcolor{lightgreen}+0.27 & \cellcolor{lightgreen}+0.14 \\
blood-transfusion-service-center & \cellcolor{lightgreen}0.54 & \cellcolor{lightgreen}0.49 & \cellcolor{lightgreen}0.48 & \cellcolor{lightgreen}0.39 & \cellcolor{lightgreen}0.61 & \cellcolor{lightgreen}0.52 & \cellcolor{lightgreen}0.66 & \cellcolor{lightgreen}0.59 & \cellcolor{lightgreen}0.68 & \cellcolor{lightgreen}0.74 & \cellcolor{lightgreen}0.70 & \cellcolor{lightgreen}0.78 & \cellcolor{lightgreen}0.84 & \cellcolor{lightgreen}0.76 & --- & \cellcolor{lightgreen}0.97 & --- & 0.85 & \cellcolor{lightgreen}+0.25 & \cellcolor{lightgreen}+0.18 \\
churn & \cellcolor{lightgreen}0.21 & \cellcolor{lightgreen}0.19 & \cellcolor{lightgreen}0.29 & 0.19 & \cellcolor{lightgreen}0.56 & 0.27 & \cellcolor{lightgreen}0.62 & \cellcolor{lightgreen}0.90 & \cellcolor{lightgreen}0.69 & --- & \cellcolor{lightgreen}0.81 & 0.45 & \cellcolor{lightgreen}0.91 & \cellcolor{lightgreen}0.90 & \cellcolor{lightgreen}0.91 & \cellcolor{lightgreen}0.90 & \cellcolor{lightgreen}0.93 & 0.90 & \cellcolor{lightgreen}+0.16 & \cellcolor{lightgreen}+0.09 \\
coil2000\_insurance\_policies & \cellcolor{lightgreen}0.42 & \cellcolor{lightgreen}0.28 & \cellcolor{lightgreen}0.47 & \cellcolor{lightgreen}0.36 & \cellcolor{lightgreen}0.66 & \cellcolor{lightgreen}0.43 & \cellcolor{lightgreen}0.71 & \cellcolor{lightgreen}0.60 & \cellcolor{lightgreen}0.74 & \cellcolor{lightgreen}0.67 & \cellcolor{lightgreen}0.83 & \cellcolor{lightgreen}0.70 & \cellcolor{lightgreen}0.79 & \cellcolor{lightgreen}0.71 & \cellcolor{lightgreen}0.87 & 0.72 & \cellcolor{lightgreen}0.96 & --- & \cellcolor{lightgreen}+0.21 & \cellcolor{lightgreen}+0.11 \\
credit-g & \cellcolor{lightgreen}0.58 & \cellcolor{lightgreen}0.57 & \cellcolor{lightgreen}0.66 & \cellcolor{lightgreen}0.57 & \cellcolor{lightgreen}0.76 & \cellcolor{lightgreen}0.63 & \cellcolor{lightgreen}0.81 & \cellcolor{lightgreen}0.78 & \cellcolor{lightgreen}0.79 & \cellcolor{lightgreen}0.80 & \cellcolor{lightgreen}0.82 & \cellcolor{lightgreen}0.94 & \cellcolor{lightgreen}0.89 & \cellcolor{lightgreen}0.94 & \cellcolor{lightgreen}0.89 & \cellcolor{lightgreen}1.00 & \cellcolor{lightgreen}0.95 & \cellcolor{lightgreen}1.00 & \cellcolor{lightgreen}+0.29 & \cellcolor{lightgreen}+0.30 \\
credit\_card\_clients\_default & \cellcolor{lightgreen}0.13 & \cellcolor{lightgreen}0.11 & 0.18 & 0.17 & 0.28 & 0.28 & 0.38 & 0.39 & 0.48 & 0.47 & 0.59 & 0.55 & 0.68 & 0.65 & 0.77 & 0.70 & \cellcolor{lightgreen}0.92 & 0.77 & -0.01 & -0.05 \\
customer\_satisfaction\_in\_airline & \cellcolor{lightgreen}0.12 & 0.08 & \cellcolor{lightgreen}0.22 & 0.19 & \cellcolor{lightgreen}0.32 & 0.28 & \cellcolor{lightgreen}0.41 & 0.38 & \cellcolor{lightgreen}0.51 & 0.45 & \cellcolor{lightgreen}0.61 & 0.57 & \cellcolor{lightgreen}0.70 & 0.66 & 0.80 & 0.78 & 0.89 & 0.88 & \cellcolor{lightgreen}+0.01 & -0.03 \\
diabetes & \cellcolor{lightgreen}0.77 & \cellcolor{lightgreen}1.00 & \cellcolor{lightgreen}0.79 & \cellcolor{lightgreen}0.80 & \cellcolor{lightgreen}1.00 & \cellcolor{lightgreen}0.84 & \cellcolor{lightgreen}1.00 & \cellcolor{lightgreen}1.00 & \cellcolor{lightgreen}1.00 & \cellcolor{lightgreen}0.99 & \cellcolor{lightgreen}1.00 & \cellcolor{lightgreen}0.85 & \cellcolor{lightgreen}0.95 & \cellcolor{lightgreen}0.96 & --- & \cellcolor{lightgreen}0.92 & --- & --- & \cellcolor{lightgreen}+0.53 & \cellcolor{lightgreen}+0.47 \\
hazelnut-spread-contaminant-detection & \cellcolor{lightgreen}0.30 & \cellcolor{lightgreen}0.20 & \cellcolor{lightgreen}0.38 & \cellcolor{lightgreen}0.34 & \cellcolor{lightgreen}0.46 & \cellcolor{lightgreen}0.40 & \cellcolor{lightgreen}0.55 & \cellcolor{lightgreen}0.51 & \cellcolor{lightgreen}0.60 & \cellcolor{lightgreen}0.58 & \cellcolor{lightgreen}0.72 & \cellcolor{lightgreen}0.66 & \cellcolor{lightgreen}0.77 & \cellcolor{lightgreen}0.74 & \cellcolor{lightgreen}0.88 & \cellcolor{lightgreen}0.81 & \cellcolor{lightgreen}0.94 & \cellcolor{lightgreen}0.93 & \cellcolor{lightgreen}+0.12 & \cellcolor{lightgreen}+0.08 \\
heloc & 0.05 & \cellcolor{lightgreen}0.26 & 0.14 & \cellcolor{lightgreen}0.29 & 0.25 & \cellcolor{lightgreen}0.35 & 0.39 & \cellcolor{lightgreen}0.44 & \cellcolor{lightgreen}0.56 & \cellcolor{lightgreen}0.50 & 0.59 & 0.54 & \cellcolor{lightgreen}0.71 & 0.60 & \cellcolor{lightgreen}0.97 & 0.76 & \cellcolor{lightgreen}0.97 & 0.69 & \cellcolor{lightgreen}+0.01 & -0.01 \\
hiva\_agnostic & --- & \cellcolor{lightgreen}0.90 & --- & \cellcolor{lightgreen}0.90 & --- & \cellcolor{lightgreen}0.90 & --- & \cellcolor{lightgreen}0.90 & --- & \cellcolor{lightgreen}0.90 & --- & --- & --- & \cellcolor{lightgreen}0.90 & --- & \cellcolor{lightgreen}0.90 & 0.07 & 0.90 & -0.83 & \cellcolor{lightgreen}+0.41 \\
in\_vehicle\_coupon\_recommendation & \cellcolor{lightgreen}0.21 & \cellcolor{lightgreen}0.15 & \cellcolor{lightgreen}0.31 & \cellcolor{lightgreen}0.25 & \cellcolor{lightgreen}0.37 & \cellcolor{lightgreen}0.32 & \cellcolor{lightgreen}0.45 & \cellcolor{lightgreen}0.43 & \cellcolor{lightgreen}0.56 & \cellcolor{lightgreen}0.52 & \cellcolor{lightgreen}0.65 & \cellcolor{lightgreen}0.63 & \cellcolor{lightgreen}0.73 & \cellcolor{lightgreen}0.73 & \cellcolor{lightgreen}0.82 & \cellcolor{lightgreen}0.81 & \cellcolor{lightgreen}0.91 & 0.90 & \cellcolor{lightgreen}+0.06 & \cellcolor{lightgreen}+0.03 \\
jm1 & \cellcolor{lightgreen}0.22 & \cellcolor{lightgreen}0.17 & \cellcolor{lightgreen}0.26 & \cellcolor{lightgreen}0.24 & \cellcolor{lightgreen}0.32 & \cellcolor{lightgreen}0.30 & \cellcolor{lightgreen}0.43 & \cellcolor{lightgreen}0.42 & \cellcolor{lightgreen}0.55 & 0.50 & \cellcolor{lightgreen}0.63 & 0.55 & \cellcolor{lightgreen}0.73 & 0.68 & \cellcolor{lightgreen}0.81 & 0.79 & \cellcolor{lightgreen}0.91 & 0.90 & \cellcolor{lightgreen}+0.04 & \cellcolor{lightgreen}+0.01 \\
kddcup09\_appetency & \cellcolor{lightgreen}0.15 & \cellcolor{lightgreen}0.16 & \cellcolor{lightgreen}0.27 & \cellcolor{lightgreen}0.22 & \cellcolor{lightgreen}0.30 & 0.24 & 0.37 & 0.38 & 0.38 & 0.38 & \cellcolor{lightgreen}0.63 & \cellcolor{lightgreen}0.66 & \cellcolor{lightgreen}0.71 & 0.70 & 0.76 & \cellcolor{lightgreen}0.91 & 0.87 & 0.61 & -0.00 & -0.03 \\
maternal\_health\_risk & \cellcolor{lightgreen}0.12 & \cellcolor{lightgreen}0.14 & \cellcolor{lightgreen}0.23 & 0.20 & \cellcolor{lightgreen}0.33 & 0.26 & \cellcolor{lightgreen}0.42 & 0.37 & \cellcolor{lightgreen}0.59 & 0.46 & \cellcolor{lightgreen}0.62 & 0.58 & \cellcolor{lightgreen}0.72 & 0.69 & \cellcolor{lightgreen}0.80 & 0.78 & \cellcolor{lightgreen}0.91 & 0.89 & \cellcolor{lightgreen}+0.03 & -0.02 \\
online\_shoppers\_intention & \cellcolor{lightgreen}0.42 & \cellcolor{lightgreen}0.35 & \cellcolor{lightgreen}0.50 & \cellcolor{lightgreen}0.47 & \cellcolor{lightgreen}0.58 & \cellcolor{lightgreen}0.47 & \cellcolor{lightgreen}0.72 & \cellcolor{lightgreen}0.56 & \cellcolor{lightgreen}0.79 & \cellcolor{lightgreen}0.58 & \cellcolor{lightgreen}0.81 & \cellcolor{lightgreen}0.73 & \cellcolor{lightgreen}0.81 & \cellcolor{lightgreen}0.82 & \cellcolor{lightgreen}0.89 & \cellcolor{lightgreen}0.85 & \cellcolor{lightgreen}0.92 & \cellcolor{lightgreen}0.96 & \cellcolor{lightgreen}+0.21 & \cellcolor{lightgreen}+0.14 \\
polish\_companies\_bankruptcy & \cellcolor{lightgreen}0.12 & \cellcolor{lightgreen}0.10 & \cellcolor{lightgreen}0.23 & 0.19 & 0.29 & 0.28 & 0.38 & 0.37 & 0.47 & 0.49 & 0.57 & 0.56 & \cellcolor{lightgreen}0.71 & 0.68 & 0.68 & \cellcolor{lightgreen}0.85 & 0.87 & 0.87 & -0.02 & -0.01 \\
qsar-biodeg & \cellcolor{lightgreen}0.50 & \cellcolor{lightgreen}0.59 & \cellcolor{lightgreen}0.62 & \cellcolor{lightgreen}0.71 & \cellcolor{lightgreen}0.67 & \cellcolor{lightgreen}0.72 & \cellcolor{lightgreen}0.72 & \cellcolor{lightgreen}0.84 & \cellcolor{lightgreen}0.77 & \cellcolor{lightgreen}0.88 & \cellcolor{lightgreen}0.82 & \cellcolor{lightgreen}0.90 & \cellcolor{lightgreen}0.88 & \cellcolor{lightgreen}0.91 & \cellcolor{lightgreen}0.91 & \cellcolor{lightgreen}0.94 & \cellcolor{lightgreen}1.00 & \cellcolor{lightgreen}0.99 & \cellcolor{lightgreen}+0.27 & \cellcolor{lightgreen}+0.33 \\
seismic-bumps & \cellcolor{lightgreen}0.72 & \cellcolor{lightgreen}0.71 & \cellcolor{lightgreen}0.74 & \cellcolor{lightgreen}0.51 & \cellcolor{lightgreen}0.71 & \cellcolor{lightgreen}0.73 & \cellcolor{lightgreen}0.71 & \cellcolor{lightgreen}0.79 & \cellcolor{lightgreen}0.75 & \cellcolor{lightgreen}0.83 & \cellcolor{lightgreen}0.77 & \cellcolor{lightgreen}0.74 & \cellcolor{lightgreen}0.79 & \cellcolor{lightgreen}0.88 & \cellcolor{lightgreen}0.99 & \cellcolor{lightgreen}0.85 & \cellcolor{lightgreen}0.92 & 0.79 & \cellcolor{lightgreen}+0.29 & \cellcolor{lightgreen}+0.26 \\
splice & \cellcolor{lightgreen}0.24 & \cellcolor{lightgreen}0.18 & \cellcolor{lightgreen}0.48 & \cellcolor{lightgreen}0.30 & \cellcolor{lightgreen}0.55 & \cellcolor{lightgreen}0.48 & \cellcolor{lightgreen}0.64 & \cellcolor{lightgreen}0.69 & \cellcolor{lightgreen}0.68 & \cellcolor{lightgreen}0.75 & \cellcolor{lightgreen}0.76 & \cellcolor{lightgreen}0.75 & \cellcolor{lightgreen}0.83 & \cellcolor{lightgreen}0.78 & \cellcolor{lightgreen}0.91 & \cellcolor{lightgreen}0.92 & \cellcolor{lightgreen}0.95 & \cellcolor{lightgreen}0.95 & \cellcolor{lightgreen}+0.17 & \cellcolor{lightgreen}+0.14 \\
students\_dropout\_and\_academic\_success & \cellcolor{lightgreen}0.29 & \cellcolor{lightgreen}0.30 & \cellcolor{lightgreen}0.41 & \cellcolor{lightgreen}0.35 & \cellcolor{lightgreen}0.49 & \cellcolor{lightgreen}0.42 & \cellcolor{lightgreen}0.56 & \cellcolor{lightgreen}0.50 & \cellcolor{lightgreen}0.61 & \cellcolor{lightgreen}0.63 & \cellcolor{lightgreen}0.66 & \cellcolor{lightgreen}0.66 & \cellcolor{lightgreen}0.73 & \cellcolor{lightgreen}0.78 & \cellcolor{lightgreen}0.83 & \cellcolor{lightgreen}0.83 & 0.89 & \cellcolor{lightgreen}0.91 & \cellcolor{lightgreen}+0.11 & \cellcolor{lightgreen}+0.10 \\
taiwanese\_bankruptcy\_prediction & \cellcolor{lightgreen}0.31 & \cellcolor{lightgreen}0.26 & \cellcolor{lightgreen}0.34 & \cellcolor{lightgreen}0.35 & \cellcolor{lightgreen}0.39 & \cellcolor{lightgreen}0.39 & \cellcolor{lightgreen}0.46 & \cellcolor{lightgreen}0.50 & \cellcolor{lightgreen}0.55 & \cellcolor{lightgreen}0.62 & \cellcolor{lightgreen}0.62 & 0.53 & \cellcolor{lightgreen}0.75 & 0.65 & \cellcolor{lightgreen}0.86 & 0.79 & 0.87 & 0.78 & \cellcolor{lightgreen}+0.07 & \cellcolor{lightgreen}+0.04 \\
website\_phishing & \cellcolor{lightgreen}0.29 & \cellcolor{lightgreen}0.48 & \cellcolor{lightgreen}0.43 & \cellcolor{lightgreen}0.59 & \cellcolor{lightgreen}0.53 & \cellcolor{lightgreen}0.76 & \cellcolor{lightgreen}0.63 & \cellcolor{lightgreen}0.89 & \cellcolor{lightgreen}0.75 & \cellcolor{lightgreen}0.92 & \cellcolor{lightgreen}0.86 & \cellcolor{lightgreen}0.93 & \cellcolor{lightgreen}0.93 & \cellcolor{lightgreen}0.97 & \cellcolor{lightgreen}0.95 & \cellcolor{lightgreen}0.98 & \cellcolor{lightgreen}0.97 & \cellcolor{lightgreen}0.99 & \cellcolor{lightgreen}+0.20 & \cellcolor{lightgreen}+0.33 \\
\midrule
\textbf{Average} & \cellcolor{lightgreen}\textbf{0.32} & \cellcolor{lightgreen}\textbf{0.33} & \cellcolor{lightgreen}\textbf{0.41} & \cellcolor{lightgreen}\textbf{0.38} & \cellcolor{lightgreen}\textbf{0.51} & \cellcolor{lightgreen}\textbf{0.46} & \cellcolor{lightgreen}\textbf{0.58} & \cellcolor{lightgreen}\textbf{0.58} & \cellcolor{lightgreen}\textbf{0.64} & \cellcolor{lightgreen}\textbf{0.64} & \cellcolor{lightgreen}\textbf{0.71} & \cellcolor{lightgreen}\textbf{0.68} & \cellcolor{lightgreen}\textbf{0.79} & \cellcolor{lightgreen}\textbf{0.77} & \cellcolor{lightgreen}\textbf{0.86} & \cellcolor{lightgreen}\textbf{0.84} & \textbf{0.90} & \textbf{0.89} & \cellcolor{lightgreen}\textbf{+0.12} & \cellcolor{lightgreen}\textbf{+0.12} \\
\bottomrule
\end{tabular}
\end{center}
\end{table}
\end{landscape}

\end{document}